\documentclass[runningheads]{llncs}
\usepackage{graphicx}
\usepackage{comment}
\usepackage{amsmath,amssymb} 
\usepackage{color}
 \usepackage{graphicx}

\begin{document}
\pagestyle{headings}
\mainmatter
\def\ECCVSubNumber{2500}  

\title{RelationGAN: When Relation Networks Meet GANs: Relation GANs with Triplet Loss} 

\titlerunning{RelationGAN}
%
\author{Runmin Wu\inst{1,2}\and
Kunyao Zhang\inst{1} \and
Lijun Wang\inst{1}\and
Yue Wang\inst{1}\and
Pingping Zhang\inst{1}\and
Huchuan Lu\inst{1}\and
Yizhou Yu\inst{2}
}
\authorrunning{R.Wu K.Zhang et al.}
%
\institute{Dalian University of Technology\and
The University of Hong Kong}
\maketitle

\begin{abstract}
Though recent research has achieved remarkable progress in generating realistic images with generative adversarial networks (GANs), the lack of training stability is still a lingering concern of most GANs, especially on high-resolution inputs and complex datasets. Since the randomly generated distribution can hardly overlap with the real distribution, training GANs often suffers from the gradient vanishing problem. A number of approaches have been proposed to address this issue by constraining the discriminator's capabilities using empirical techniques, like weight clipping, gradient penalty, spectral normalization etc. In this paper, we provide a more principled approach as an alternative solution to this issue. Instead of training the discriminator to distinguish real and fake input samples, we investigate the relationship between paired samples by training the discriminator to separate paired samples from the same distribution and those from different distributions. To this end, we explore a relation network architecture for the discriminator and design a triplet loss which is better generalization and stability.
Extensive experiments on benchmark datasets show that the proposed relation discriminator and new loss can provide significant improvement on variable vision tasks including unconditional and conditional image generation and image translation.
\keywords{Generative Adversarial Network, Triplet Loss}
\end{abstract}

\section{Introduction}
Since first proposed in~\cite{GAN14}, generative adversarial networks (GANs) have a rapid development and numerous applications in many computer vision tasks, such as image generation~\cite{GAN14,relativistic,sagan}, person re-identification~\cite{FD-GAN}, image super-resolution~\cite{super-resolution} etc.
It also has been extended to natural language processing~\cite{nlp}, video sequence synthesis~\cite{video}, and speech synthesis~\cite{speech}.

Though tremendous success has been achieved, training GANs is still a very tricky process and suffers from many issues, including the instability between the generator and the discriminator as well as the extremely subtle sensitivity to network and hyper-parameters. Most of these issues are due to the fact, that the support of both target distribution and generated distribution are often of low dimension regarding to the base space, and therefore misaligned at most of the time, causing discriminator to collapse to a function that hardly provides gradients to the generator. 

To remedy this issue, recent works proposed to leverage the Integral Probability Metric (IPM), such as Gradient Penalty~\cite{wgangp} and Spectral Normalization~\cite{SNGAN}. In IPM-based GANs, the discriminator is constrained to a specific class of function so that it does not grow too quickly and thus alleviates vanishing gradients. 

However, the existing IPM methods also have their limits. For instance, the hyperparameter tuning of gradient penalty is mostly empirical, while the spectral normalization imposes constrains on every which hinders the learning capacity of discriminators.  

In~\cite{relativistic}, the authors argue that non-IPM-based GANs are missing a relativistic discriminator, which IPM-based GANs already possess. The relativistic discriminator is necessary to make the training process analogous to divergence minimization and produce sensible predictions based on the prior knowledge that half of the samples in the mini-batch are fake. 
Although they have shown the power of relativistic discriminator, the potential of comparing the relation between real and fake distribution still remains to be explored.

In this paper, we explicitly study the effect of relation comparison in GANs by training the discriminator to determine whether the input paired samples are drawn from the same distribution (either real or fake). A relation network is present, acting as the discriminator. A new triplet loss is also designed for training the GANs. In this way, the problem of the disjointed support could be alleviated by projecting and merging the low dimension data distribution into a high dimension feature space. 
\begin{figure*}[t]
		\includegraphics[width=0.95\linewidth, height=0.25\linewidth]
		{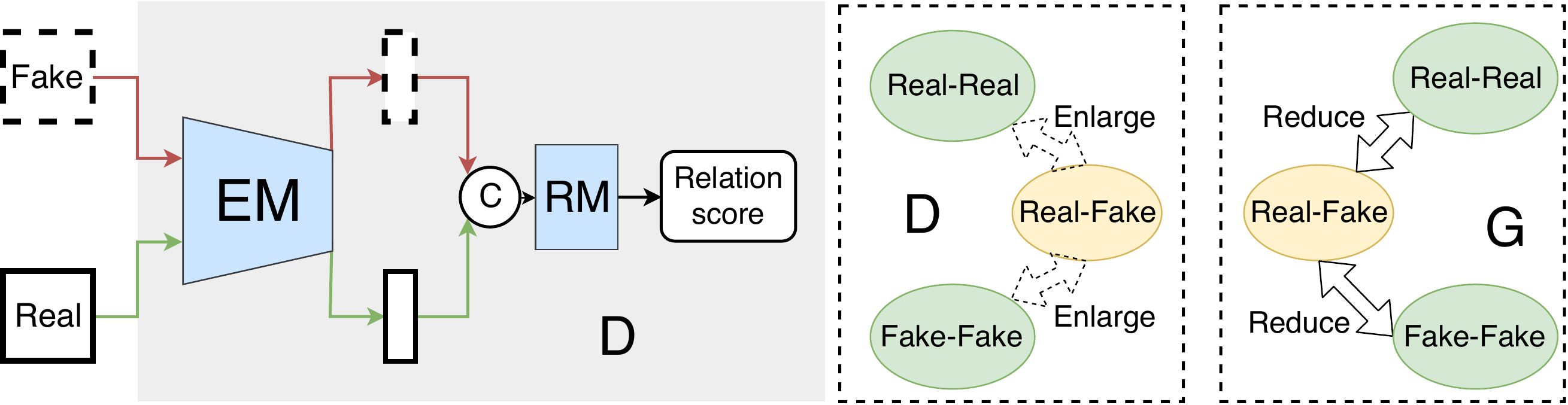} 
	\caption{Illustration of our Relation GAN, the ``C'' denotes contactenation operation. Our discriminator consists of two modules, an embedding module (EM) and a relation module (RM). The relation discriminator is expected to enlarge the difference of relation score between asymmetric pairs and symmetric pairs while the generator is expected to reduce the difference.}	
    \label{fig:General}
    \vspace{-2mm}
\end{figure*}
Mathematically, we analyze the oscillatory behavior that GANs exhibit for the Dirac-GAN and we demonstrate the proposed Relation GAN is locally convergent even with no regularized methods. Extensive experiments are conducted on  conditional and unconditional image generation and image translation tasks. The promising performance demonstrates the proposed relation gan has great potential in various applications of GANs.


In summary, the contributions of this paper are two folds.
\begin{itemize}
\item We propose a new training strategy for GANs to better leverage the relation between samples. Instead of separating real samples from generated ones, the discriminator is trained to determine whether a paired samples are from the same distribution.
\item We propose a relation network as the discriminator and a triplet loss for training GANs. We show both theoretically and empirically that the relation network together with the triplet loss give rise to generated density which can exactly match that of real data. 
  

\end{itemize}
Extensive experiments on 2D grid~\cite{unroll}, Stacked MNIST~\cite{MNIST}, CelebA~\cite{CelebA}, LSUN~\cite{lsun}, CelebA-HQ~\cite{celebahq} data sets confirm our proposed method performs favourably against state-of-the-arts such as relativistic GAN~\cite{relativistic}, WGAN-GP~\cite{wgangp}, Least Squares GAN (LSGAN)~\cite{LSGAN} and vanilla GAN~\cite{GAN14}.
\section{Related Work}
The vanilla GAN~\cite{GAN14} minimizes the Jensen–Shannon(JS) divergence of two distributions, leading to the gradient vanishing problem when the two distributions are disjoint. Recent works try to address this issue by designing new objective functions~\cite{LSGAN,LS-GAN,ganqp,wgan} or more sophisticated network architectures~\cite{progressive,sagan,biggan,dcgan}. Others investigate the regularization and/or normalization to constrain the ability of discriminator~\cite{SNGAN,wgangp,dragan}. Recently, a new method~\cite{relativistic} is proposed to explore a relativistic discriminator. 
In the following, we will review recent works using different objective functions and a special case--relativistic GANs, which are closely related to our approaches. 
\subsection{Different Objective Functions in GANs}
Generally, there are two kind of loss functions in GANs: the minimax GAN and the non-saturating (NS) GAN.
In the former the discriminator minimizes the negative loglikelihood for the binary classification task. In the latter the
generator maximizes the probability of generated samples
being real. The non-saturating loss
as it is known to outperform the minimax variant empirically. Among them, loss sensitive GAN~\cite{LS-GAN} tries to solve the problem of gradient vanishing by focusing on training samples with low authenticity.
WGAN~\cite{wgan} proposes the Wasserstein distance to replace the JS divergence, which can measure their distance even though the two distributions are disjoint. In addition, 
\cite{wgan} also proposes to add noise to both real and generated samples to further alleviate the impact of disjoint distributions. \cite{wgangp} improves WGAN by replacing the weight clipping constraints with a gradient penalty, which enforces the Lipschitz constraint on the discriminator by punishing the norm of the gradient.
DRAGAN~\cite{dragan} combines the two parts of WGAN and LSGAN, and only improves the loss function to a certain extent. The stability of loss training is controlled by constantly updating the coefficient of the latter term.

\subsection{Relativistic GANs}
Instead of training discriminators to predict the absolute probabilities of the input samples being real, the relativistic GAN~\cite{relativistic} proposes to use a relativistic discriminator, which estimates the probability of the given real sample being more realistic than a randomly sampled fake sample. 
Although bears a similar spirit, our method differs from \cite{relativistic} in that we adopt a relation network as the discriminator to estimate the relation score of a paired input. In comparison, the discriminator in \cite{relativistic} treats input samples separately and
relies on a ranking loss (\emph{e.g.}, hinge loss) to explore their relation. The idea of merging the features and comparing the relation between samples from two distribution has not been explored in the literature of GANs. In addition, our method proposes a new triplet loss to leverage the power of paired relation comparison, allowing more stability and better diversity for GANs without applying any IPM methods.


\section{The Relation GAN Framework}
\subsection{Relation Net Architecture}
In traditional GANs, a discriminator is trained to distinguish real samples from fake ones and a generator is trained to confuse the discriminator by generating realistic samples.
Consider a real data distribution $P_{data}$, and the data distribution $P_{G^*}$ produced by the generator $G$.
Rather than training the discriminator on real and fake data independently, we propose to train a discriminator which predicts a relation score $D(x^1,x^2)$ for a paired input, indicating whether the paired samples are from the same distribution (either $P_{data}$ or $P_{G^*}$).

Inspired by the success of relation net architecture in other computer vision areas~\cite{relationnet}, our discriminator consists of two modules, including an embedding module and a relation module as shown in Figure~\ref{fig:General}. For a pair of input samples, the embedding module firstly maps each sample into a high dimensional feature space. Their corresponding features are then merged and fed into the relation module to produce the relation score for the input pair. For ease of description, we name paired inputs containing both real and fake samples as \textbf{asymmetric pairs}, and those containing samples from the same distribution (either real or fake) as \textbf{symmetric pairs}. The embedding module is jointly trained with the relation module.
The training process is then formulated as a min-max game (See Section~\ref{sec:min-max}), where the discriminator aims to maximize the relation scores of asymmetric sample pairs and minimize those of symmetric ones. Meanwhile, the generator is trained to confuse the discriminator by minimizing the relation scores of asymmetric sample pairs containing real and generated samples.


\subsection{The Min-Max Game}\label{sec:min-max}
The min-max game in training GANs is conducted by optimizing the losses of $D$ and $G$ iteratively.
In the no-IPM GANs, the generalized losses of $D$ and $G$ can be presented as follows:
\begin{equation} \label{eq1}
{L_{D}=\mathbb{E}_{x \sim \mathbb{P}_{data},y\sim \mathbb{P}_{G^*}}[\tilde{f}(D(x),D(y))]}
\end{equation}
and 
\begin{equation} \label{eq2}
L_{G}=\mathbb{E}_{x\sim \mathbb{P}_{data},y\sim \mathbb{P}_{G^*}} [\tilde{g}(D(x),D(y))] 
\end{equation} 
where $\tilde{f}$ and $\tilde{g}$ are scalar-to-scalar functions, \(\mathbb{P}_{data}\) is the distribution of real data, and \(P_{G^*}\) denotes the generated data distribution. 

Let \(\tilde{f}(x,y)=-log(x)-log(1-x)\), and \(\tilde{g}(x,y)=-log(y)\). Eq \eqref{eq1} and \eqref{eq2} will become the loss functions of the standard GAN~\cite{GAN14}.
$y$ from $P_{G^*}$ and $x$ from $P_{data}$.

In our Relation GAN, the formulation of the losses functions of $D$ and $G$ are as follows:

\begin{align}
L_{D}=\mathbb{E}_{x_{r}\sim \mathbb{P}_{data}, {y}_{f}\sim \mathbb{P}_{G^*}}[\tilde{f}(D(x_{r}^1,x_{r}^2),D(y_{f}^1,y_{f}^2),D(x_{r}^1,y_{f}^1))]
\end{align}
and
\begin{equation}
L_{G}= \mathbb{E}_{x_{r}\sim \mathbb{P}_{data}, {y}_{f}\sim \mathbb{P}_{G^*}}[\tilde{g}(D(x_{r}^1,x_{r}^2), D(y_{f}^1,y_{f}^2),D (x_{r}^1,y_{f}^1))]
\end{equation}
where $\tilde{f}$ and $\tilde{g}$ are also scalar to scalar function. 

The goal of relation discriminator is to learn a loss function $D_{\theta}$ parameterized by $\theta$ which separates symmetric and asymmetric sample pairs by a desired margin. Then the generator can be trained to minimize this margin by generating realistic samples.

Inspired by the success of triplet loss~\cite{siamese_trip}, we formulate the similar loss function in our Relation GAN as follows:
\begin{align}
L_D=&\mathbb{E}_{{x_{r} \sim \mathbb{P}_{data}},{y_{f} \sim \mathbb{P}_{G^*}}}[(D(x_r^1,x_r^2)-D(x_r^1,y_f^1)+\Delta(x_r^2,y_f^1)]_{+}\notag\\
+&\mathbb{E}_{{x_{r}\sim\mathbb{P}_{data}},{y_{f} \sim \mathbb{P}_{G^*}}}[(D(y_f^1,y_f^2)-D(x_r^1,y_f^1)+\Delta(y_f^2,x_r^1)]_{+}
\end{align}

and 
\begin{equation}\label{ours}
\begin{aligned}
L_G=&\mathbb{E}_{{x_{r} \sim \mathbb{P}_{data}},{y_{f} \sim \mathbb{P}_{G^*}}}[D(x_r^1,y_f^1)]-\mathbb{E}_{{x_{r} \sim \mathbb{P}_{data}}}[D(x_r^1, x_r^2)]\notag\\
&+\mathbb{E}_{{x_{r} \sim \mathbb{P}_{data}},{y_{f} \sim\mathbb{P}_{G^*}}}[D(x_r^1,y_f^1)]-\mathbb{E}_{{y_{f} \sim \mathbb{P}_{G^*}}}[D(y_f^1, y_f^2)]\notag\\
=&2\mathbb{E}_{{x_{r} \sim \mathbb{P}_{data}},{y_{f} \sim \mathbb{P}_{G^*}}}[D(x_r^1,y_f^1)]-\mathbb{E}_{{y_{f} \sim \mathbb{P}_{G^*}}}[D(y_f^1, y_f^2)]
\end{aligned}
\end{equation}
where  $x_r^1$ and $x_r^2$ are samples from the real data distribution, $y_f^1$ is sample from the generated data distribution and $y_f^2$ is sample from the data generated by the generator in the last step of optimization. We use a distance metric to replace the constant `margin' in the original triplet loss.  This variable constraining leads to a smaller difference of relation scores when the distance between the two compared samples are smaller, which is more flexible than the original fixed margin. Our experiments also shows the superiority of our new triplet loss with $\Delta$ margin.

\subsection{A Variant Loss}\label{rmmr}
Since the training batch size is limited, the sampled distribution of each batch may deviates from the real data distribution. For an input batch of $m$ paired samples, the loss function in \eqref{ours} can be written as follows:

\begin{align}\label{relu-mean}
L_D=&\frac{1}{m}\sum\nolimits_{i=1}^{m} [(D(x_r^{1i},x_r^{2i})-D(x_r^{1i},y_f^{1i})+\Delta(x_r^{2i}y_f^{1i})]_{+}\notag\\
+&\frac{1}{m}\sum\nolimits_{i=1}^{m}[(D(y_f^{1i}, y_f^{2i})-D(x_r^{1i},y_f^{1i})+\Delta(y_f^{2i},x_r^{1i})]_{+}
\end{align}
where ${x_{r} \sim \mathbb{P}_{data}},{y_{f} \sim \mathbb{P}_{G^*}}$. Our triplet loss is designed to reduce the relation scores of symmetric sample pairs and increase those of asymmetric ones.

However, when the real sample distribution is fairly uniform with small variance, the original loss is rigorous and prone to be disturbed by outliers in one batch. For these cases, we design a variant of our new triplet loss as follows:
\begin{align}\label{mean-relu}
L_D=& [(\frac{1}{m}\sum\nolimits_{i=1}^{m}D(x_r^{1i},x_r^{2i})-\frac{1}{m}\sum\nolimits_{i=1}^{m}D(x_r^{1i},y_f^{1i})\notag
\\& +\frac{1}{m}\sum\nolimits_{i=1}^{m}\Delta(x_r^{2i}y_f^{1i})]_{+}\notag\\
+&[\frac{1}{m}\sum\nolimits_{i=1}^{m}(D(y_f^{1i}, y_f^{2i})-\frac{1}{m}\sum\nolimits_{i=1}^{m}D(x_r^{1i},y_f^{1i})\notag\\&
+\frac{1}{m}\sum\nolimits_{i=1}^{m}\Delta(y_f^{2i},x_r^{1i})]_{+}
\end{align}

where $i$ represents the index of samples in a batch.  
The variant loss is more relaxed and not easily disturbed by the extreme samples in the same batch. It performs better on evenly distributed data sets.

Thus, we suggest to employ the variant triplet loss on uniform distribution data, e.g., datasets with only single class data. Our experiments results on the dataset of single class such as, CelebA and LSUN confirm it.

\section{Theory Proof and Analysis}
As discussed in the introduction, the optimal discriminator of most GANs is a divergence. In this section, we firstly prove that the proposed discriminator based on the relation net also has such property, and then show the distributional consistency under our Lipschitz continuous assumption. 

\subsection{A New Divergence}
A divergence is a function $\mathcal{M}$ of two variables $p$, $q$ satisfies the following
definition:

\textbf{Definition 1} If $\mathcal{M}$ is function of two variables $p$, $q$ satisfies the following properties:

1. \(\mathcal{M}[p, q] \geq 0\)

2. \(p=q \Leftrightarrow \mathcal{M}[p, q]=0\)

Then $\mathcal{M}$ is a divergence between $p$ and $q$.

\textbf{Assumption 1}
In the training process, when $G$ not reach the optimal $G^*$, $y_f^1$ ought to be more realistic than $y_f^2$, and $D$ ought to give bigger relation score to the paired input $(y_f^1,y_f^2)$ than $(x_r^1,x_r^2)$. $y_f^1$ ought to be more realistic than $y_f^2$ also means,  $\Delta(y_f^2,x_r)$ is bigger than $\Delta(y_r^1,x_r)$

Under this assumption, we show the loss function of our relation  discriminator $L_{D}$ is also a divergence as follows.
We demonstrate the $L_{D}$ of equation(8) is a divergence and show that the $L_{D}$ satisfies the two requirements in \textbf{Definition 1}.

First, 
\begin{align}
&L_D=\notag\\
&\mathbb{E}_{{x_{r} \sim \mathbb{P}_{data}},{y_{f} \sim \mathbb{P}_{G^*}}}[(D(x_r^1,x_r^2)-D(x_r^1,y_f^1)
+\Delta(x_r^2,y_f^1)]_{+}\notag\\
+&\mathbb{E}_{{x_{r}\sim\mathbb{P}_{data}},{y_{f} \sim \mathbb{P}_{G^*}}}[(D(y_f^1, y_f^2)-D(x_r^1,y_f^1)+\Delta(y_f^2,x_r^1)]_{+}\notag\\
&\geq 
0+0=0
\end{align}

Second, when $G$ and $D$ reaches the optimal state, we have $P_{G^*}=P_{data}$ and $y_f^2=y_f^1=x_r$, 
ignore the bias in the sampled batch (assume the batch-size is infinite), 
we have
\begin{align}
&L_D=\notag\\
&\mathbb{E}_{{x_{r} \sim \mathbb{P}_{data}},{y_{f} \sim \mathbb{P}_{G^*}}}[(D(x_r,x_r)-D(x_r,x_r)+\Delta(x_r,x_r)]_{+}\notag\\
+&\mathbb{E}_{{x_{r}\sim\mathbb{P}_{data}}}[(D(x_r,x_r)-D(x_r,x_r)+\Delta(x_r,x_r)]_{+}\notag
\\&=0
\end{align}

Finally, when $G$ and $D$ not reached the optimal state, we have $P_{G^*}\neq P_{data}$ and $y_f^2\neq y_f^1$ ($y_f^1$ ought to be more realistic than $y_f^2$ cause the $y_f^2$ is sample from the generated sample by the generator in last iteration). Under the \textbf{Assumption 1}, we have:
\begin{align}
&L_D=\notag
\\&\mathbb{E}_{{x_{r} \sim \mathbb{P}_{data}},{y_{f} \sim \mathbb{P}_{G^*}}}[(D(x_r^1,x_r^2)-D(x_r^1,y_f^1)+\Delta(x_r^2,y_f^1)]_{+}\notag
\\&
+\mathbb{E}_{{x_{r}\sim\mathbb{P}_{data}},{y_{f} \sim \mathbb{P}_{G^*}}}[(D(y_f^1, y_f^2)-D(x_r^1,y_f^1)+\Delta(y_f^2,x_r^1)]_{+}\notag\\
>
&\mathbb{E}_{{x_{r} \sim \mathbb{P}_{data}},{y_{f} \sim \mathbb{P}_{G^*}}}[(D(x_r^1,x_r^2)-D(x_r^1,y_f^1)+\Delta(x_r^2,y_f^1)]_{+}\notag\\
+&\mathbb{E}_{{x_{r}\sim\mathbb{P}_{data}},{y_{f} \sim \mathbb{P}_{G^*}}}[(D(x_r^1,x_r^2)-D(x_r^1,y_f^1)+\Delta(y_f^1,x_r^2)]_{+}\notag\\
&= 2 \times\mathbb{E}_{{x_{r} \sim \mathbb{P}_{data}},{y_{f} \sim \mathbb{P}_{G^*}}}[(D(x_r^1,x_r^2)-D(x_r^1,y_f^1)+
\notag\\&
\Delta(x_r^2,y_f^1)]_{+}
\geq0
\end{align}

From (1) (2) (3), we prove the $L_{D}$ satisfies all the requirements of a divergence. 

\subsection{Distributional Consistency}
We use $\theta$ to denote the parameterized function discriminator and $\phi$ to denote the parameterized function of generator. Based on~\cite{LS-GAN}, we use the definition of Lipschitz assumption of data density as follows:

\textbf{Lemma 1} Under Assumption 2, for a Nash equilibrium \(\left(\theta^{*}, \phi^{*}\right)\) in Lemma 1, we have

\(\int_{\mathbf{x}} | P_{\text {data}}(x)-P_{G^{*}}(x) | d x \leq 0\)

Thus, \(P_{G^{*}}(x)\) converges to \(P_{d a t a} (x)\). 
The proof of this lemma is given in the \textbf{Supplementary 2}.
\begin{figure*}[h]
\begin{tabular}{c@{}c@{}c@{}c@{}c@{}}
		\includegraphics[ width=0.20\linewidth,height=0.20\linewidth]
		{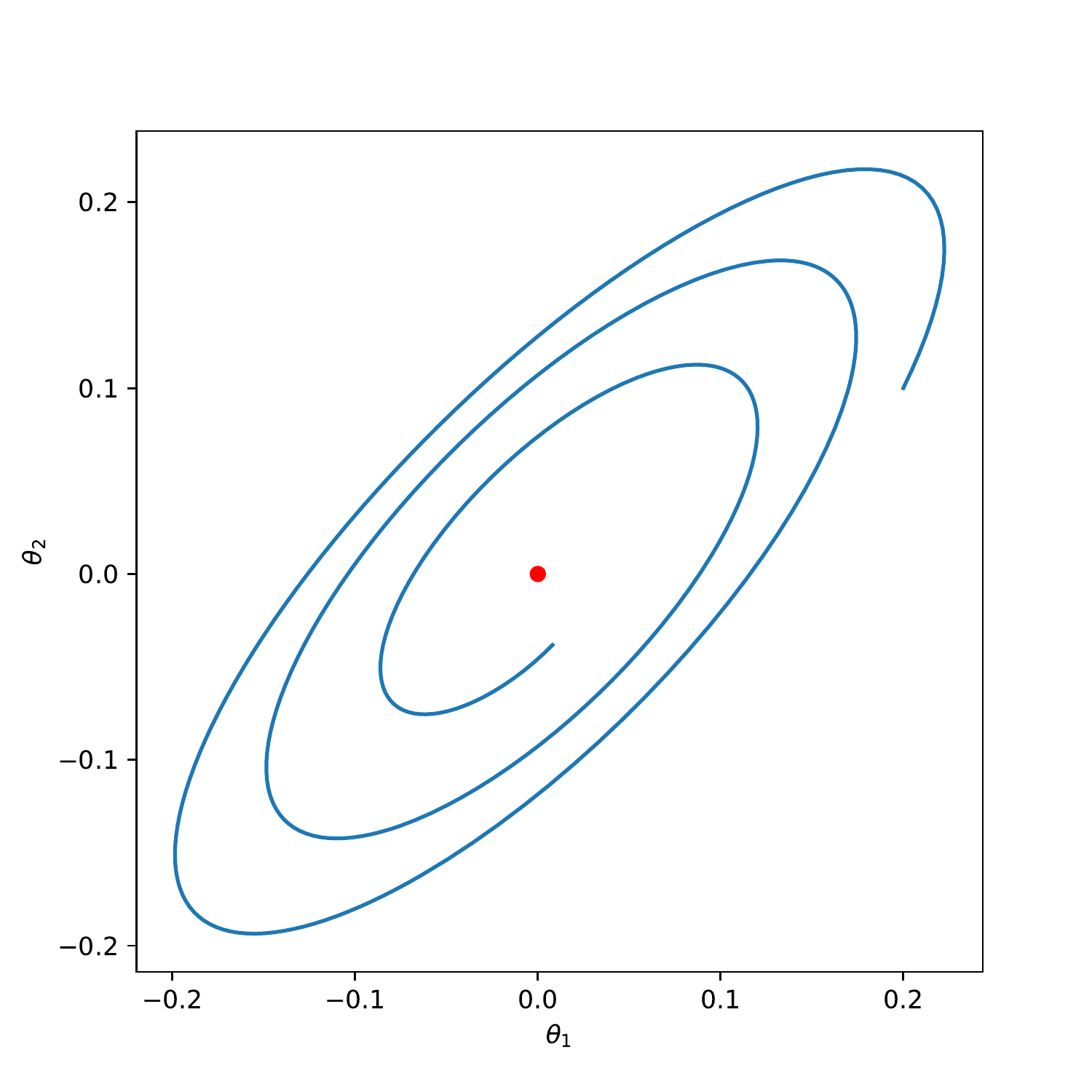} \ 
		&\includegraphics[width=0.20\linewidth, height=0.20\linewidth]
		{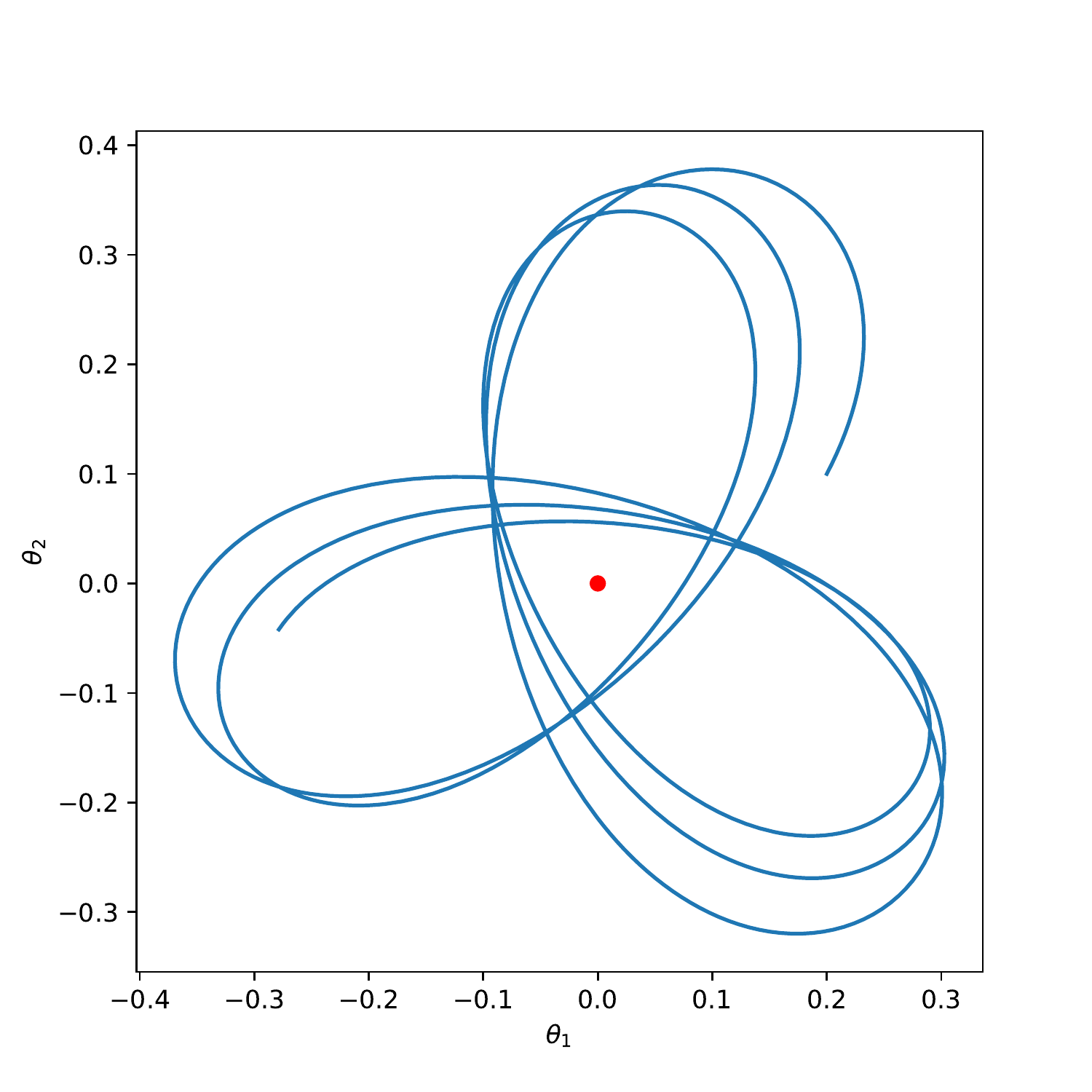}& \
		\includegraphics[width=0.20\linewidth,height=0.20\linewidth]
		{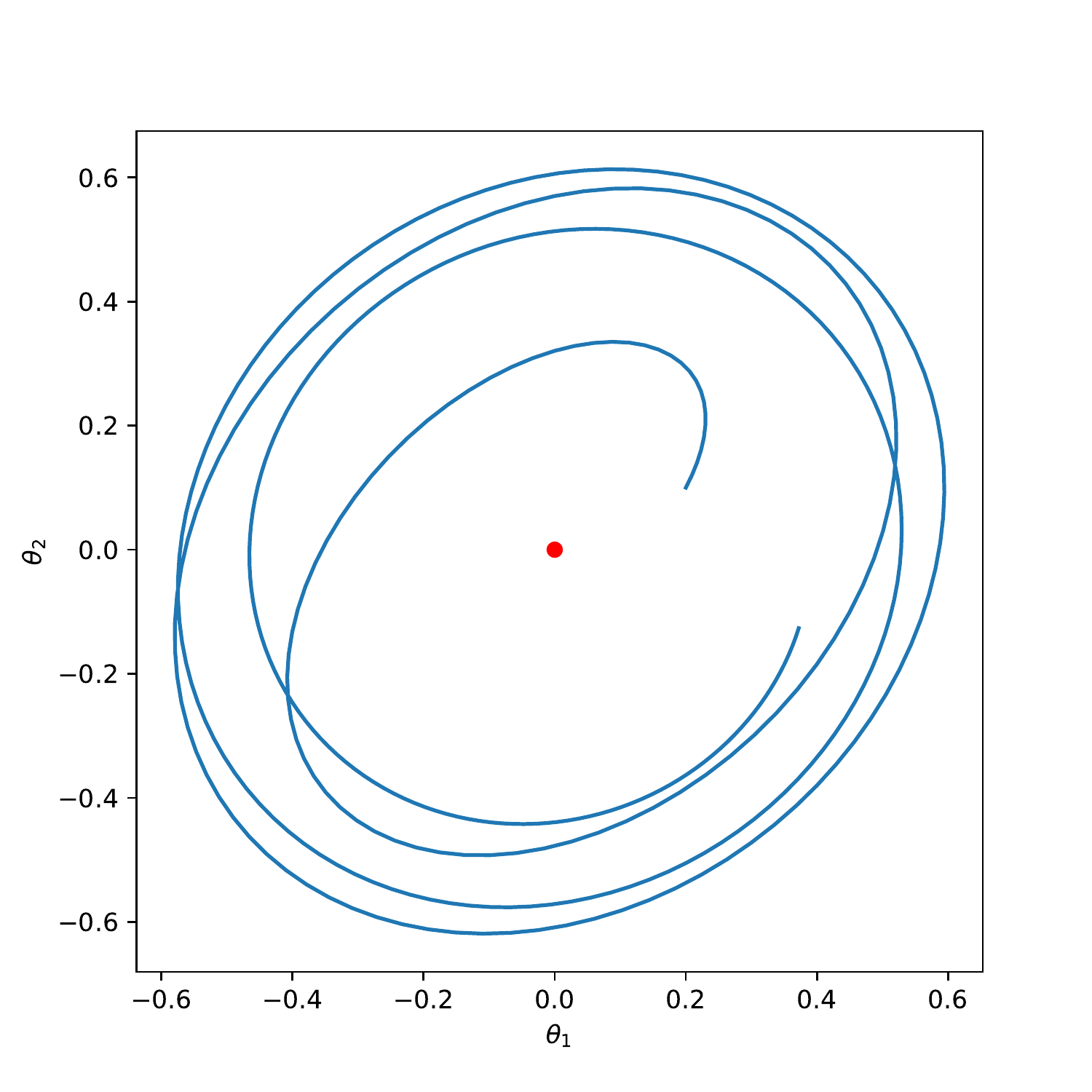} \ 
		&\includegraphics[width=0.20\linewidth,height=0.20\linewidth]
		{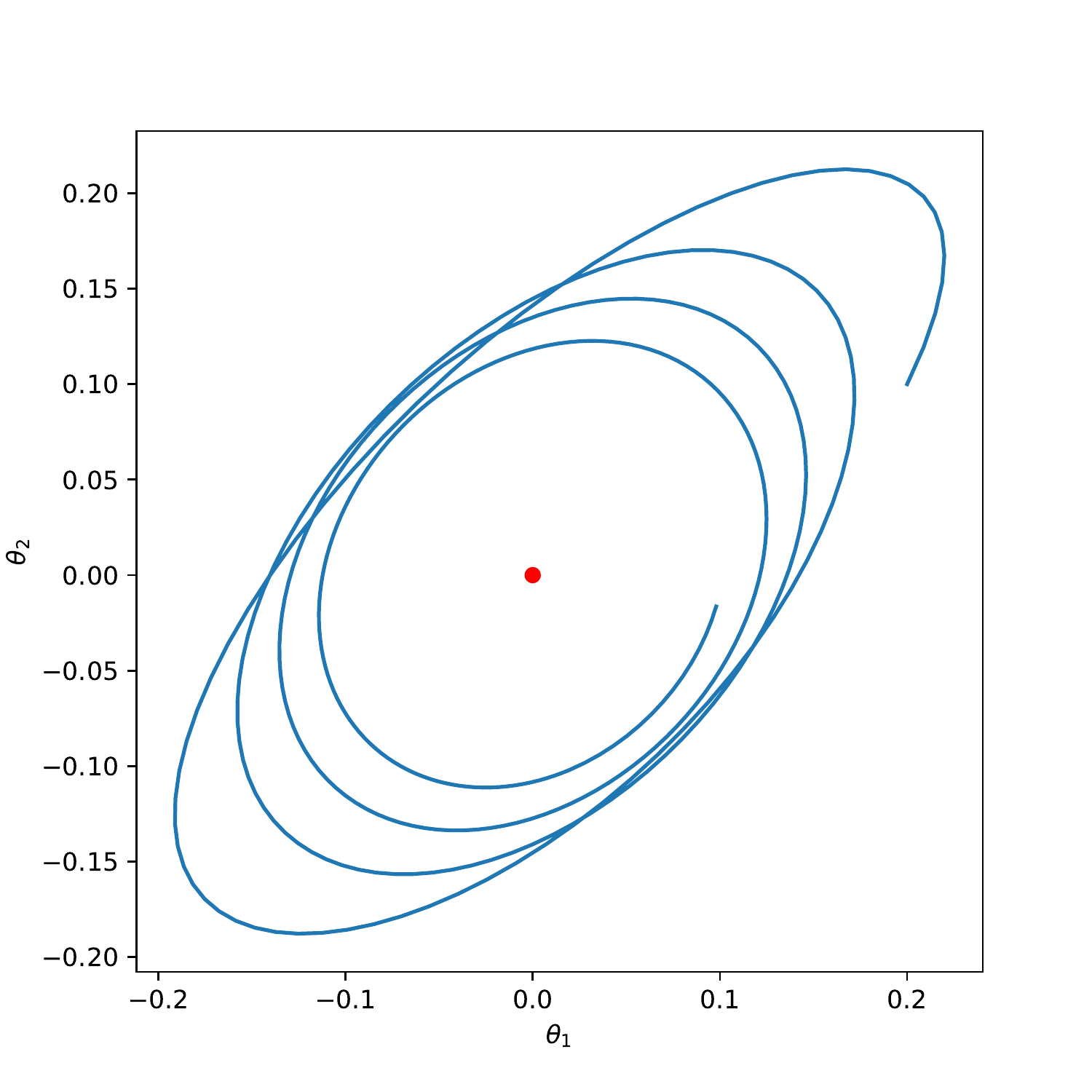} \ 
		&\includegraphics[width=0.20\linewidth, height=0.20\linewidth]
		{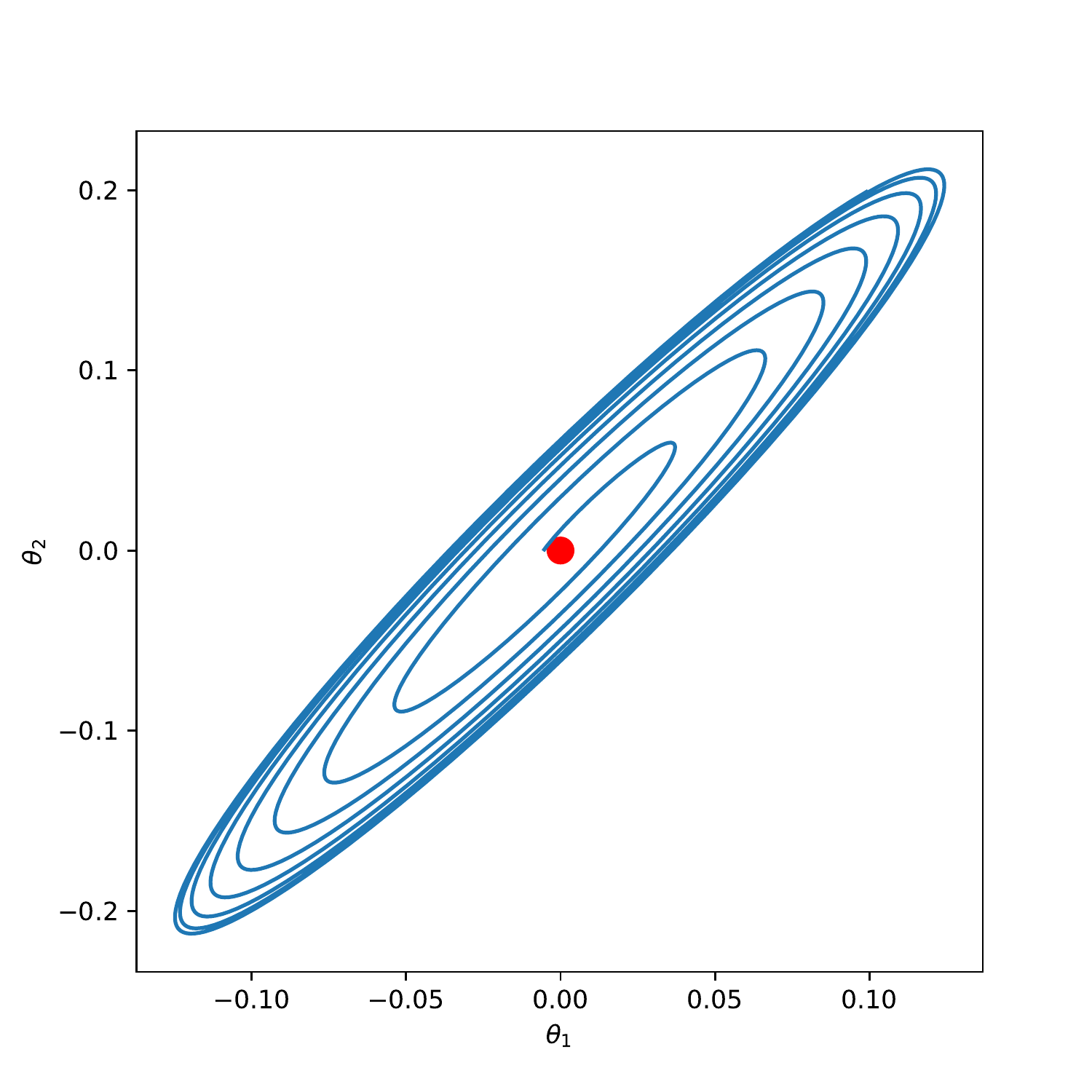}
		\\
        {\footnotesize (a) Vanilla GAN} &
		{\footnotesize (b) WGAN}
		&	{\footnotesize (c) WGAN-GP}
		&	{\footnotesize (d) GAN-QP}
		&{\footnotesize (e) Relation GAN}\\
	\end{tabular}
	\caption{The numerical solution on Dirac GANs} 		
    \label{fig:Local Stability}
    \vspace{-3mm}
\end{figure*}
\subsection{The Convergence}
In the literature, GANs are often treated as dynamic systems to study their training convergence~\cite{actuallyconverge},~\cite{numericgan},~\cite{localstable},~\cite{TTUR}.
This idea can be dated back to the Dirac GAN~\cite{actuallyconverge},
which describes a simple yet prototypical counterexample for understanding whether the GAN training is locally nor globally convergent. To further analyze the convergence rate of training the proposed Relation GAN, we also adopt the Dirac GAN theory.
However, \cite{actuallyconverge} only discusses the situation where the data distributions are 1-D. We extend this theory into the 2-D case to gain better understanding.

\textbf{Definition 3} The Dirac-GAN consists of a (univariate) generator distribution \(p_{\theta}=\delta_{\theta}\) and a linear discriminator $D(x)=\phi \cdot x $, where $\theta$ denotes the parameter of the generator, $x$ is a 2-D vector, and $\phi$ represents the parameter of the discriminator. The real data distribution $p_{data}$ is a Dirac-distribution concentrated at $(0,0)$.

 Suppose the real sample point is a vector $(0,0)$, and the fake sample is being reorganized, which also represents a parameter of the generator. The discriminator uses the simplest linear model, \emph{i.e.}, $D(x)=x$, which also represents the parameters of the discriminator. Dirac GAN takes into account that in such a minimalist model, whether a false sample eventually converges to a true sample, in other words, whether a  finally converges to $(0,0)$. Specifically, in Relation GAN, our Dirac Discriminator could be simplified as: $D(x^1,x^2)={\theta}_r({\theta}_e \cdot x^1+{\theta}_e \cdot x^2)$, where $\theta_r$ and $\theta_e$ denotes the parameter of the embedding module and relation module respectively.

Based on the dynamic analysis for GANs in \textbf{Supplementary 3}, we have the numerical solution of the GANs' dynamic equations with a initial point $(0.2,0.1)$ as the fig~\ref{fig:Local Stability} shows.
In~\cite{actuallyconverge}, the author find that most unregulared GANs are not locally convergent. In our 2-D Dirac
 GANs, the numerical solutions of the WGAN~\cite{wgan}, WGAN-GP~\cite{wgangp}, GAN-QP~\cite{ganqp}, vanilla GAN~\cite{GAN14} also perform oscillating near the real sample or hard to converge to the real sample point, while our Relation GAN success to converge. It indicates that our GAN has a good local convergence.  
 
\section{Experiments}
We first evaluate the proposed Relation GAN on the 2D synthetic dataset and the Stacked MNIST dataset to demonstrate the diversity of generated data and the stability of generator.
We then perform the image generation tasks with our method to show its superiority in synthesizing natural images.
Finally, ablation study is conducted to verify the effects of the feature merging mechanism in relation networks and the proposed triplet loss.
\small{
\begin{figure*}[h!]
\begin{tabular}{c@{}c@{}c@{}c@{}c@{}}
\\
		\includegraphics[width=0.20\linewidth,height=0.15\linewidth]
		{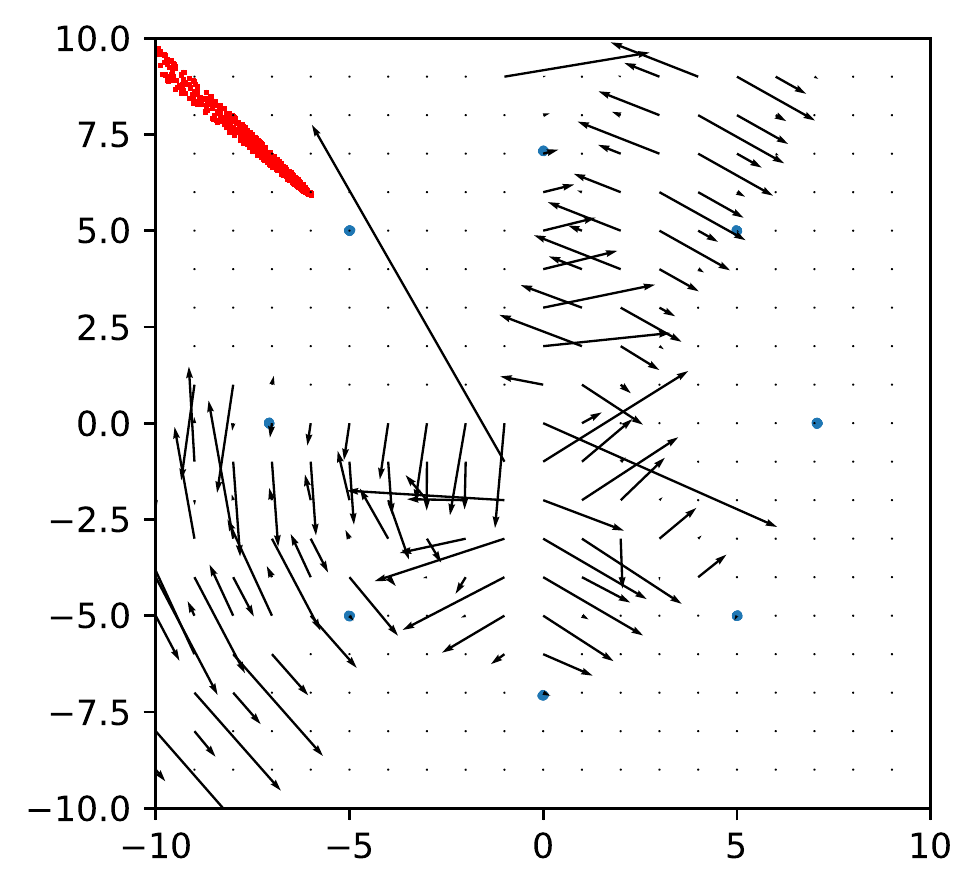}\ 
		&\includegraphics[width=0.20\linewidth,height=0.15\linewidth]
		{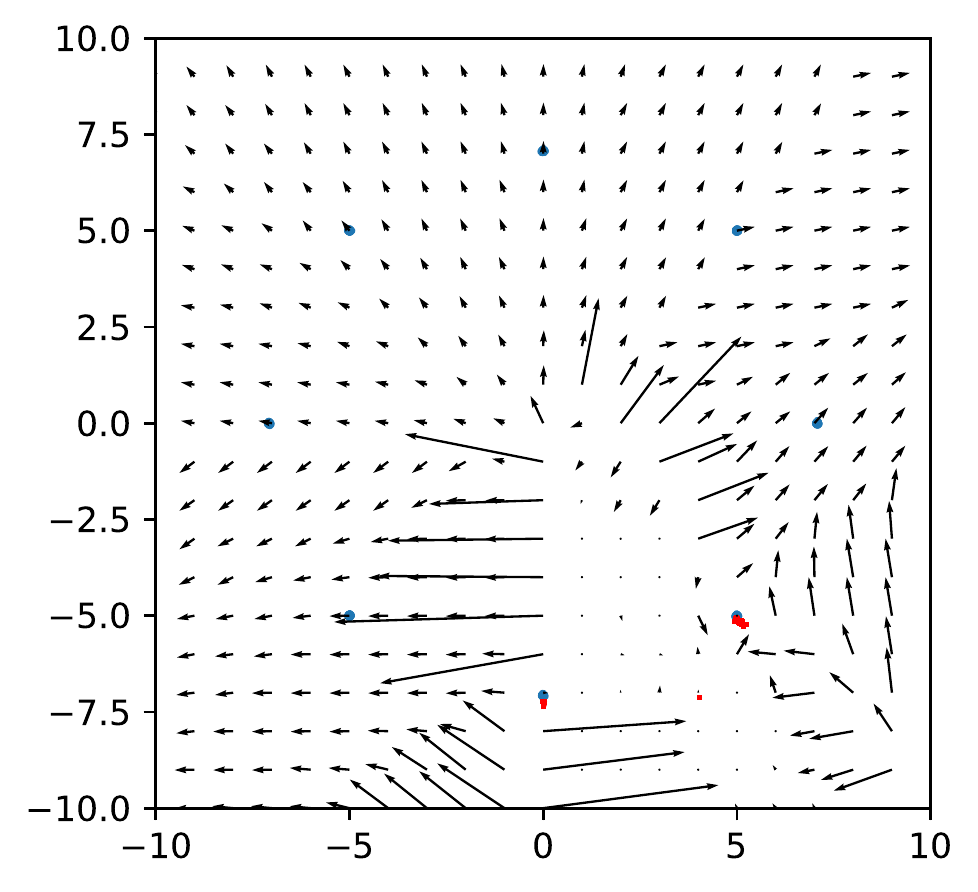}\
		&\includegraphics[width=0.20\linewidth,height=0.15\linewidth]
		{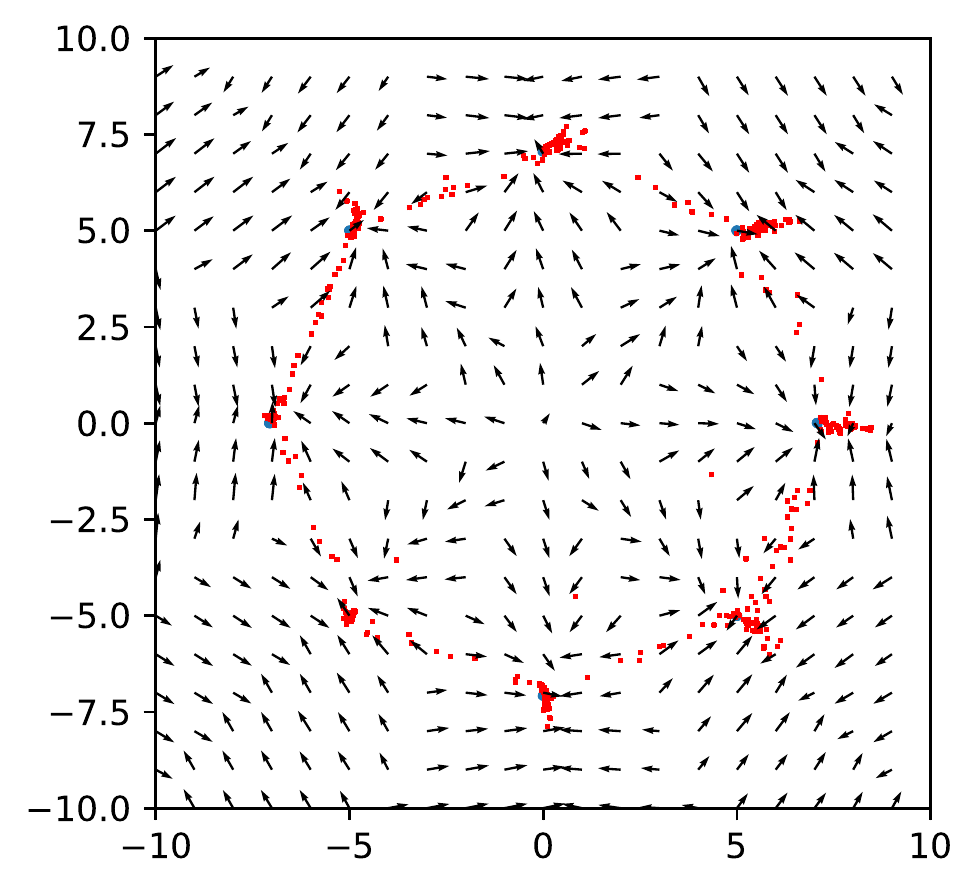}\
		 &\includegraphics[width=0.20\linewidth,height=0.15\linewidth]
		{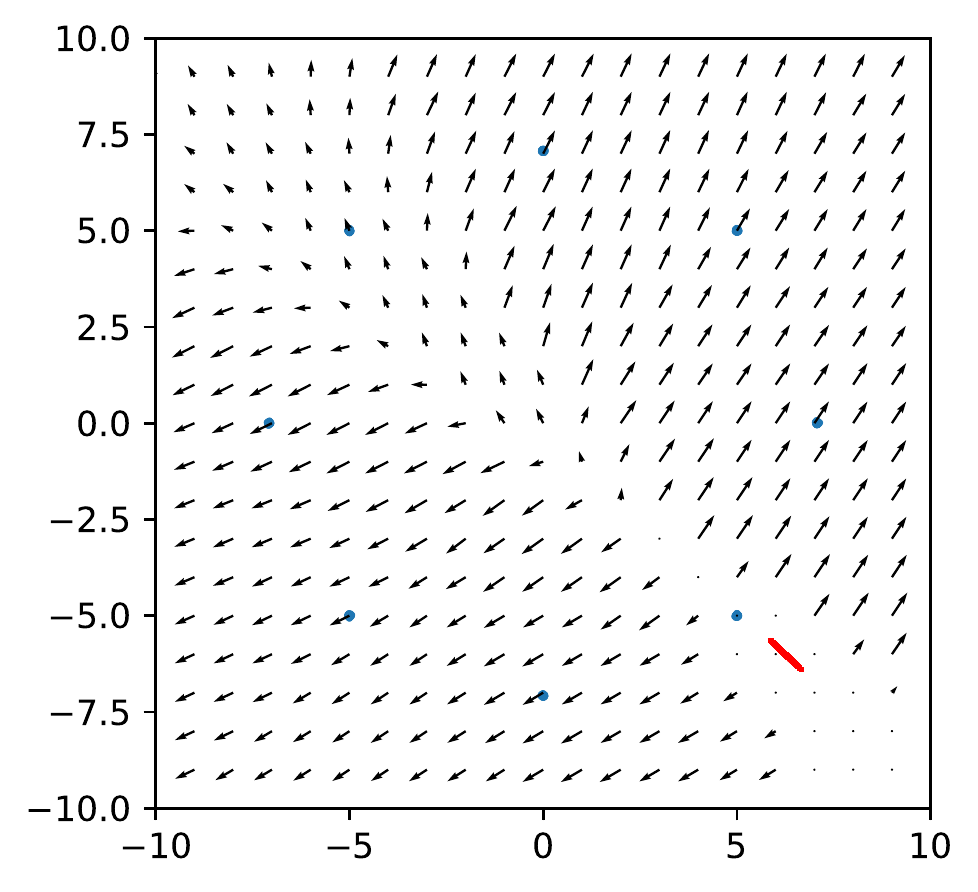}  \
		&\includegraphics[width=0.20\linewidth,height=0.15\linewidth]
		{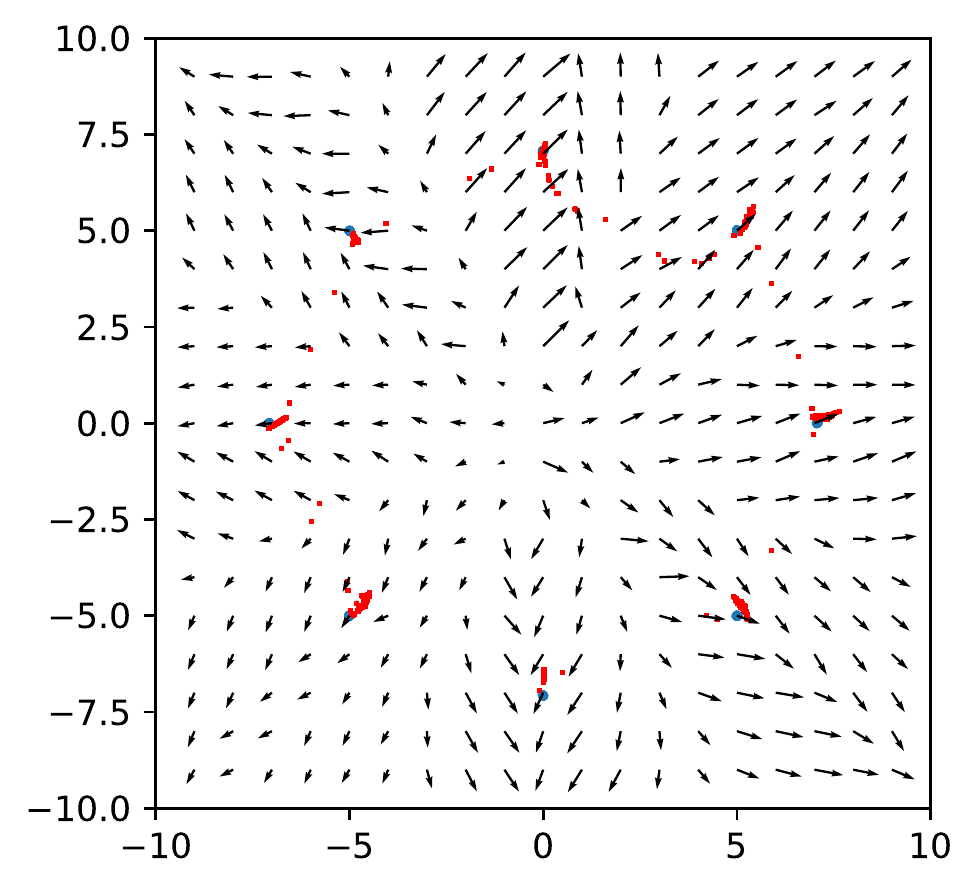}\
		\\
		\includegraphics[width=0.20\linewidth, height=0.15\linewidth]
		{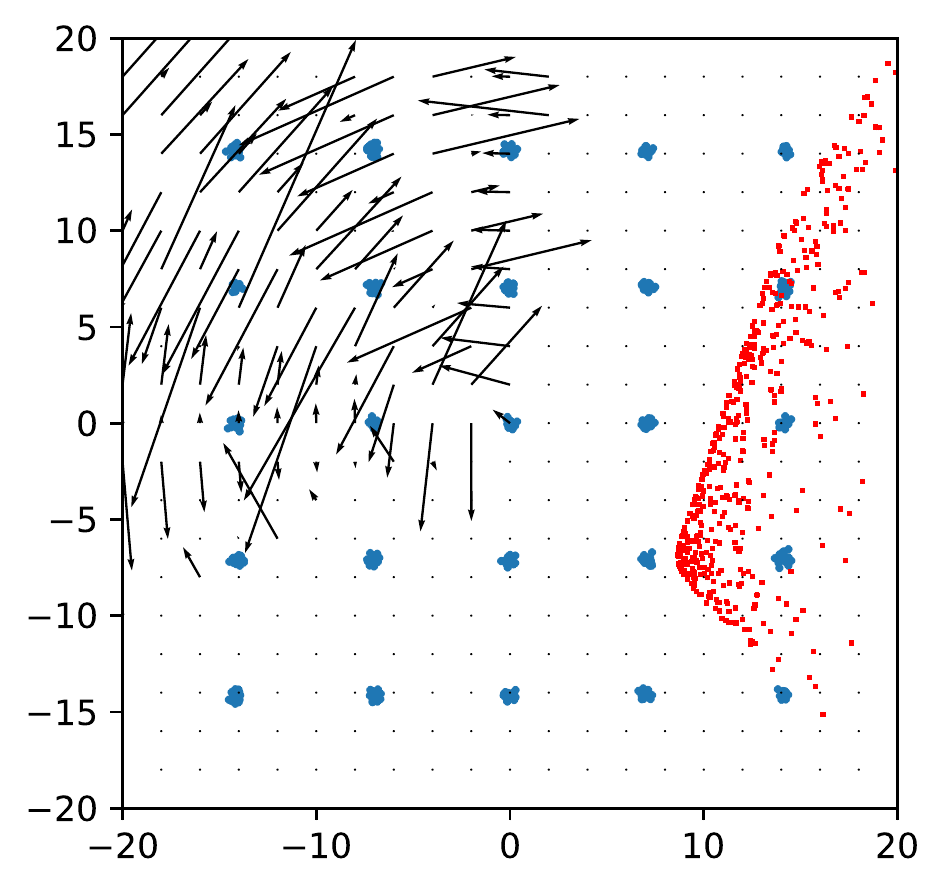}\ 
		&\includegraphics[width=0.20\linewidth, height=0.15\linewidth]
		{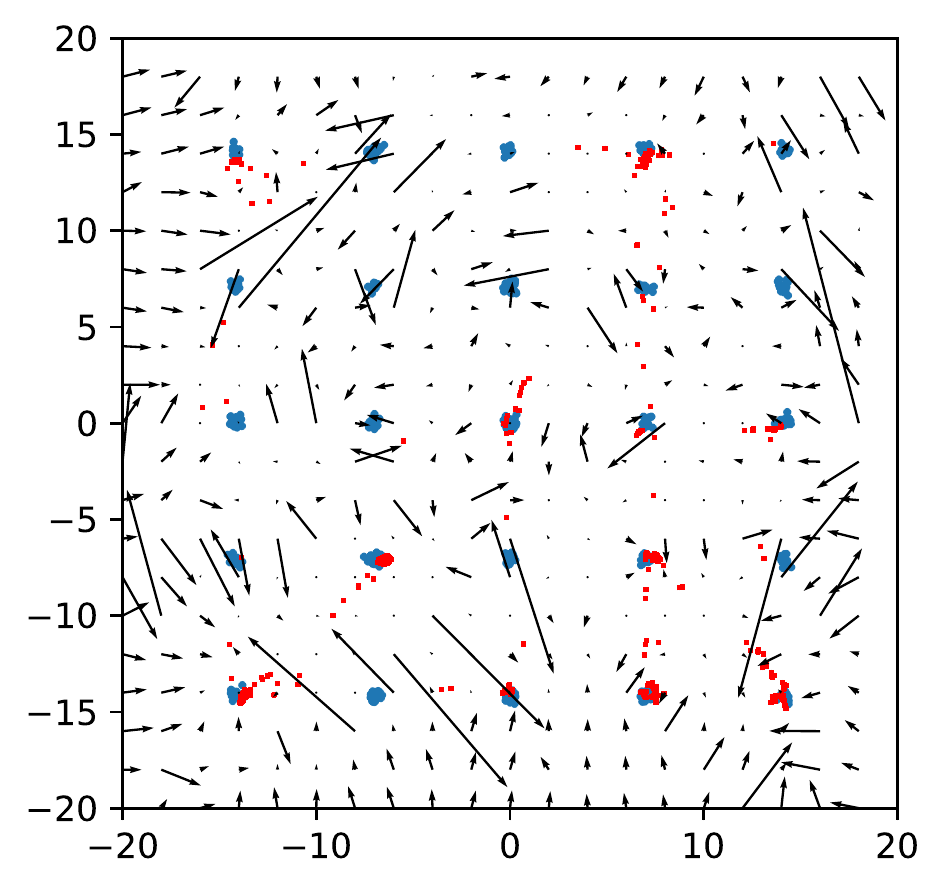}\
		&\includegraphics[width=0.20\linewidth,height=0.15\linewidth]
		{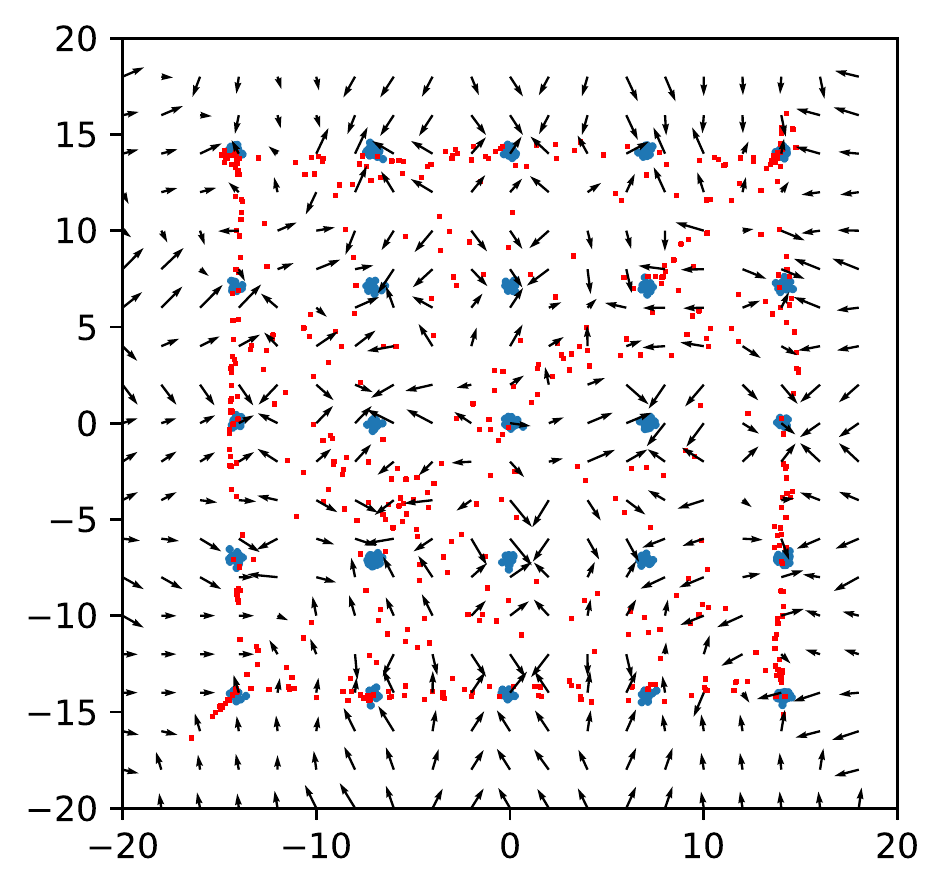}\
		&\includegraphics[width=0.20\linewidth,height=0.15\linewidth]
		{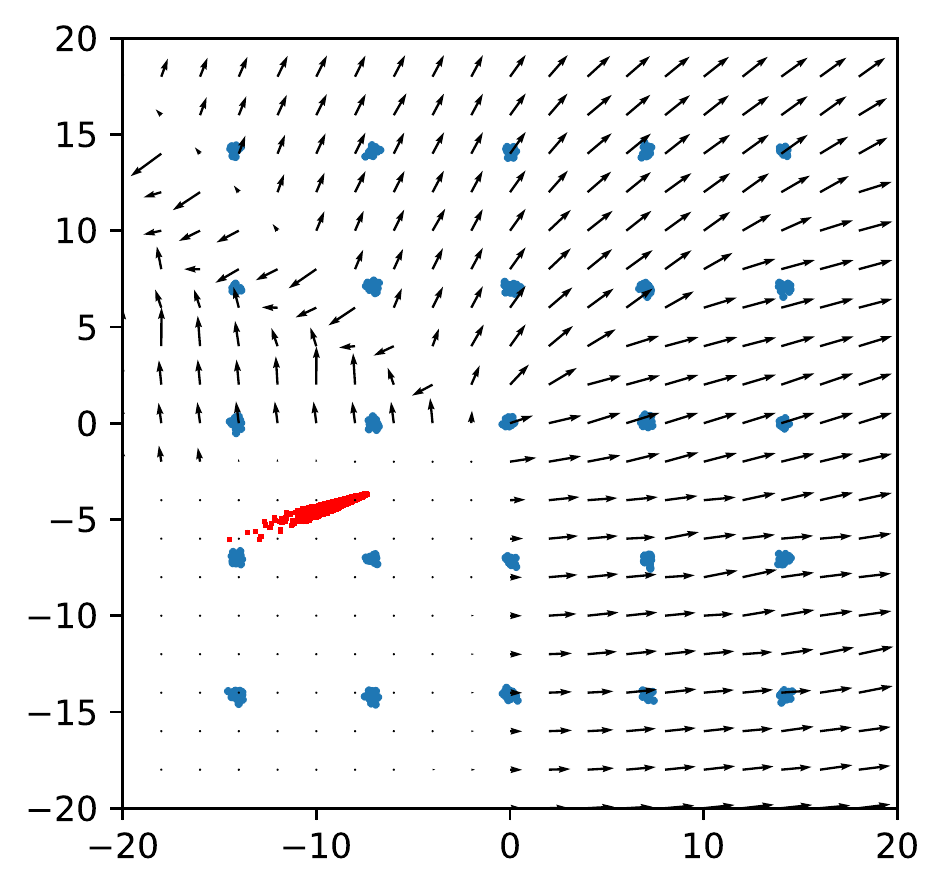}  \
		&\includegraphics[width=0.20 \linewidth,height=0.15\linewidth]
		{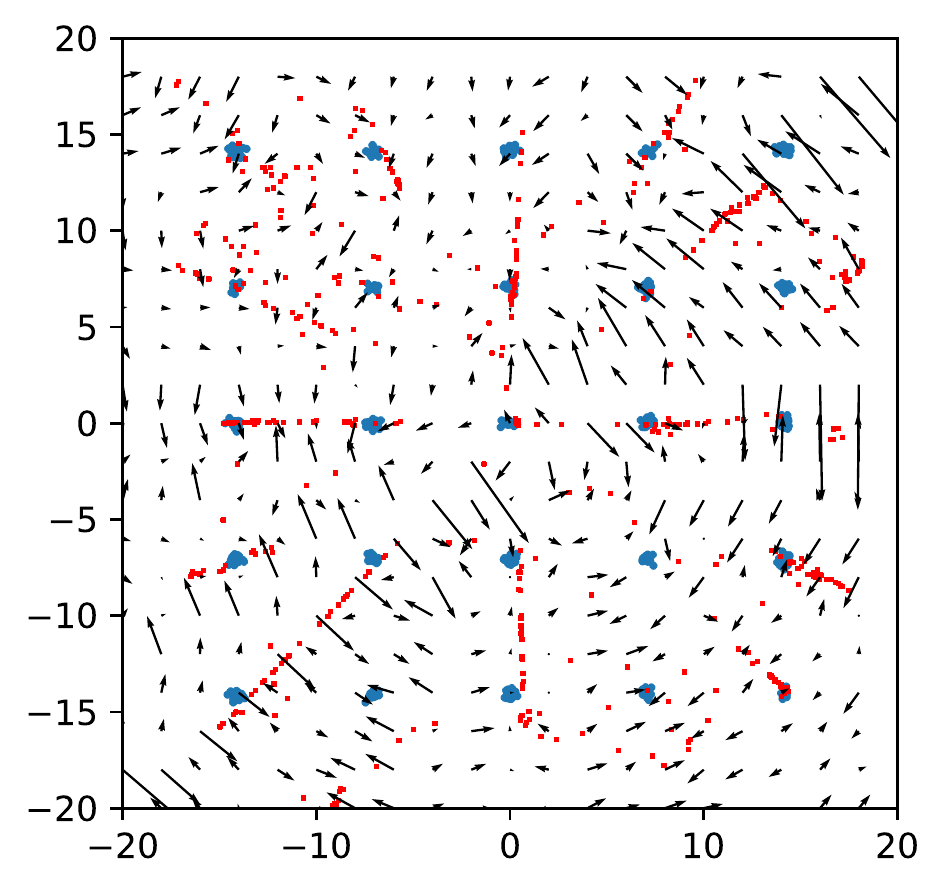}  \
		\\
		\includegraphics[width=0.20\linewidth, height=0.15\linewidth]
		{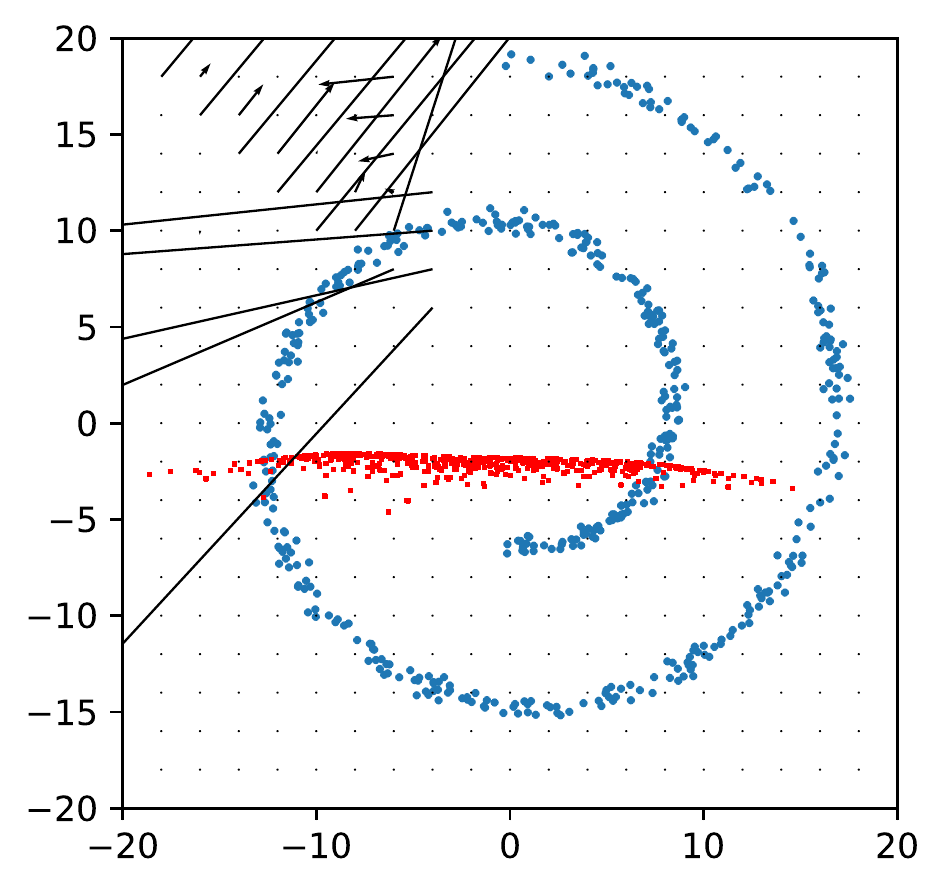} \ 
		&\includegraphics[width=0.20\linewidth, height=0.15\linewidth]
		{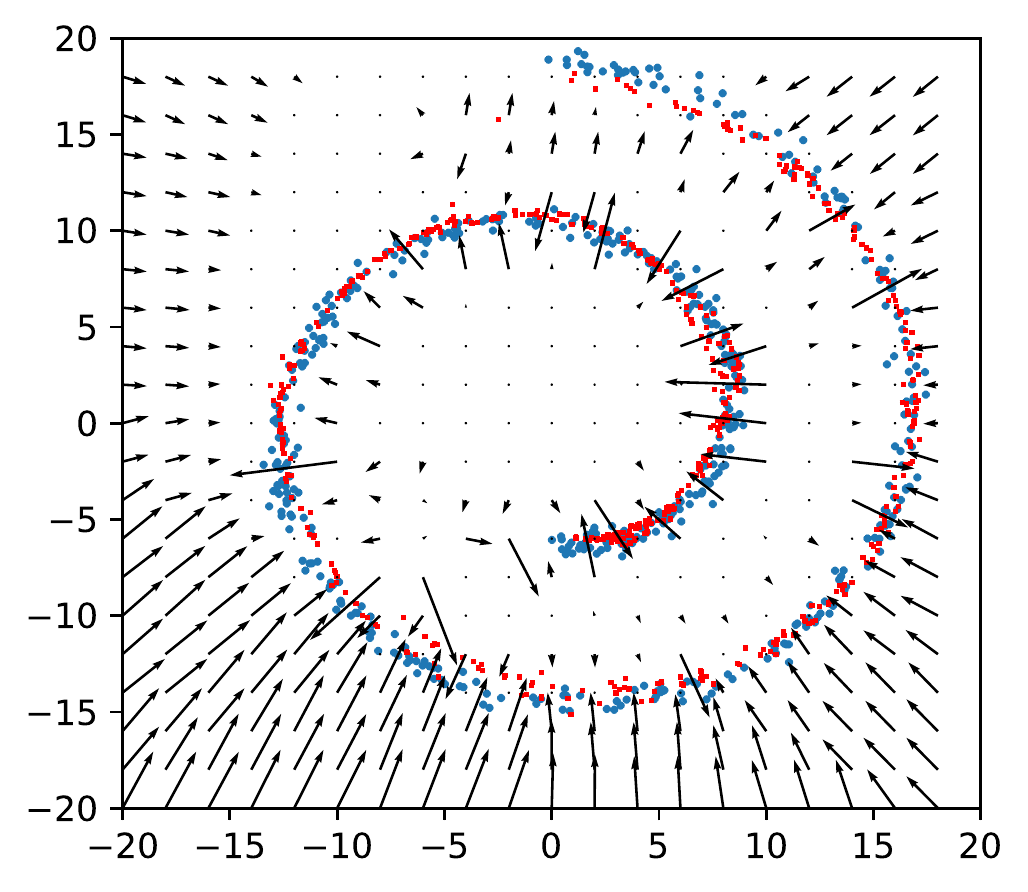}\
		&\includegraphics[width=0.20\linewidth,height=0.15\linewidth]
		{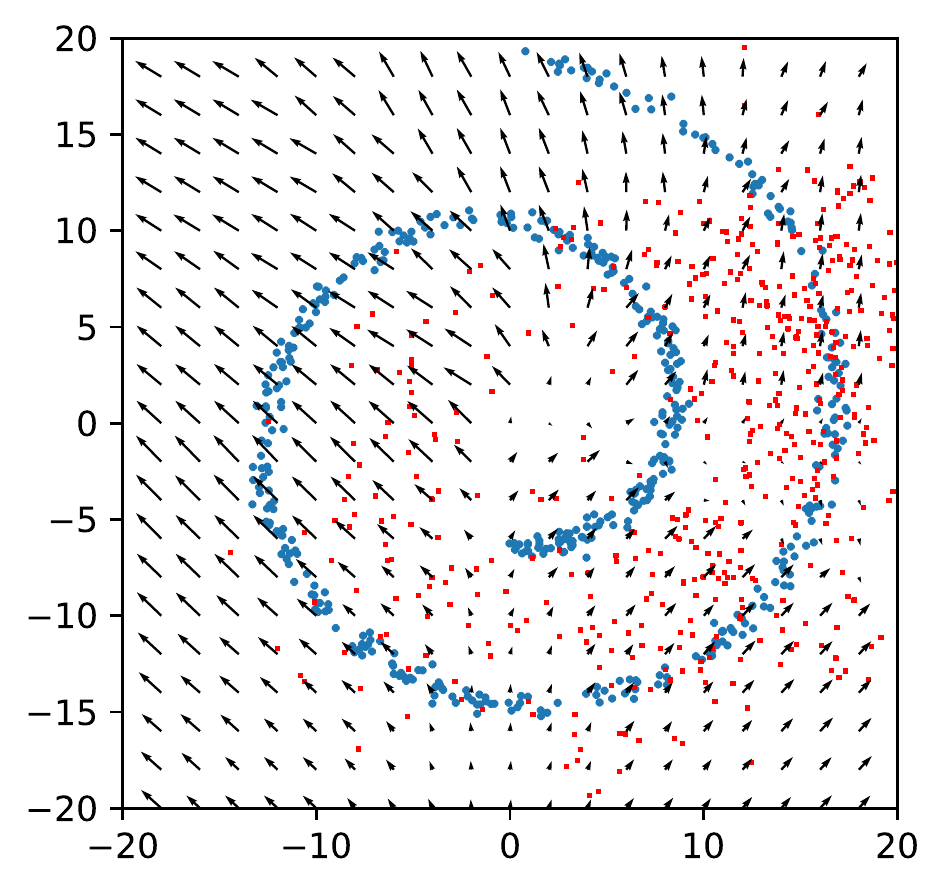}  \
		&\includegraphics[width=0.20\linewidth,height=0.15\linewidth]
		{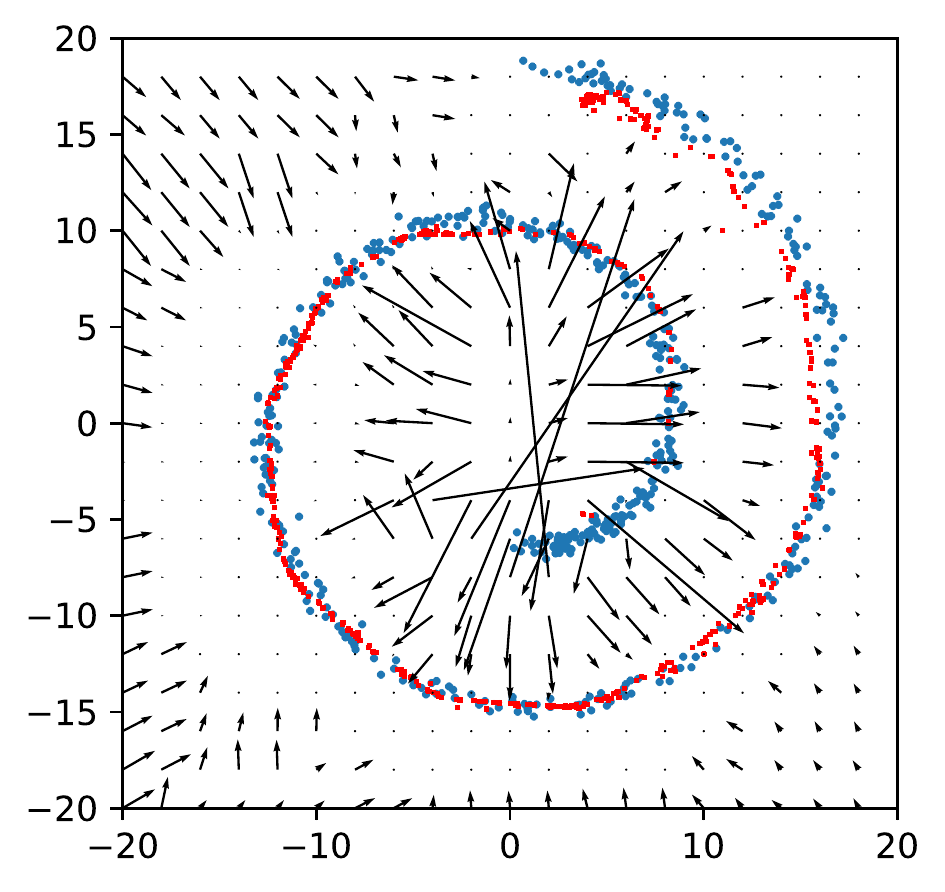}  \
		&\includegraphics[width=0.20\linewidth,height=0.15\linewidth]
		{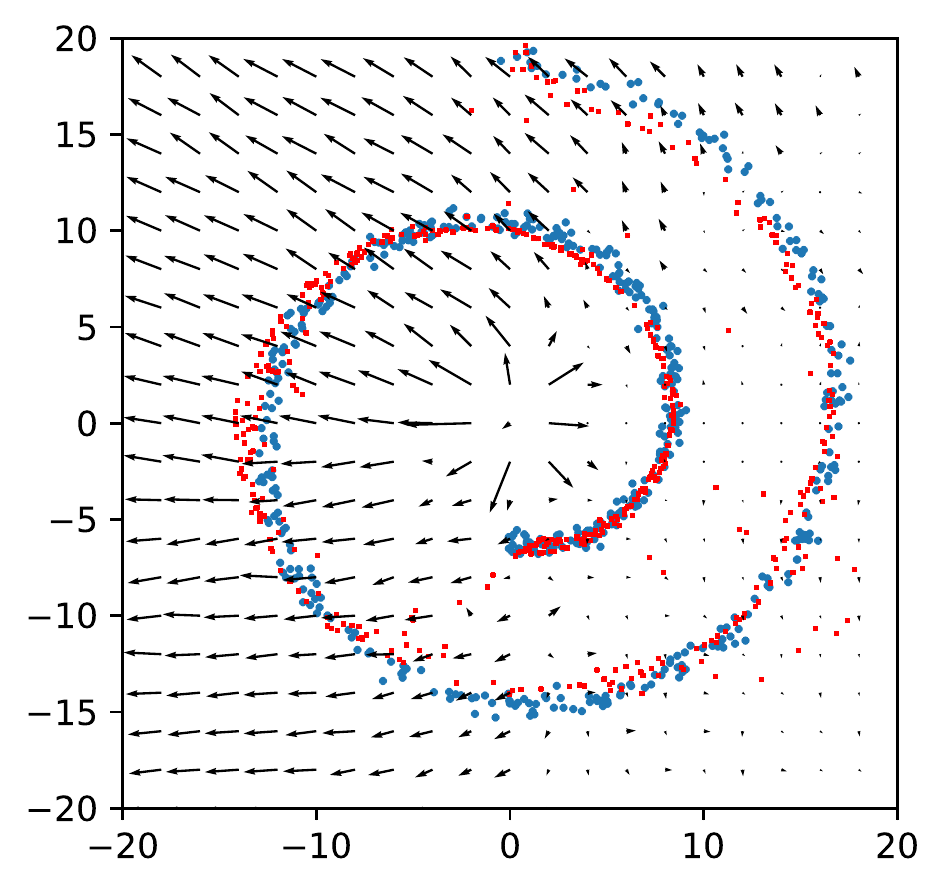}  \
		\\
        {\footnotesize (a) Vanilla GAN} &
		{\footnotesize (b) LSGAN}
		&	{\footnotesize (c) WGAN-GP }&
		{\footnotesize (d)Relativistic}&
		{\footnotesize (e) Relation GAN }
	\\
	\end{tabular}
	\caption{Comparison on 2D datasets.} 		
    \label{2D}
    \vspace{-4mm}
\end{figure*}}
\\
\subsection{The Diversity of Generated Data}
\textbf{2D Datasets}
We compare the effect of our relation discriminator on the 2D 8-Gaussian distribution, 2D 25-Gaussian distribution and 2D swissroll distribution. The experimental settings follow \cite{generationstability}. The results generated by our method and four popular methods under the same setting are shown in Figure~\ref{2D}. Compared with the other methods, ours can better fit these 2D distributions.
\\
\textbf{Stacked MNIST}
For Stacked MNIST~\cite{MNIST} experiments, we use the setting and code of ~\cite{generationstability}. Each of the three channels in each sample is classified by a pre-trained MNIST classifier, and the resulting three digits determine which of the 1000 modes the sample belongs to. We measure the number of modes captured with the pre-trained classifier. We choose  Adam~\cite{adam} optimizer for all experiments. Our results are shown in Table~\ref{MNIST}. We find that our Relation GAN could achieve best mode coverage, reaching all 1,000 modes. 
 \begin{table*}
  \centering
     \caption{Stacked MNIST}\label{MNIST}
  \begin{tabular}{ll}
  \hline
    Loss&{Modes}                   \\
    \hline
    LSGAN  & 985$\pm$10        \\
    WGANGP   & 643$\pm$7    \\
    Vanilla GAN    &  923$\pm$18     \\
    Relativastic GAN     &  828$\pm$58     \\
    Ours           &1000$\pm$0 \\
   \hline
\end{tabular}
\vspace{-4mm}
  \end{table*}

\subsection{Unconditional Image Generation}
\textbf{Datasets} We provide comparison on four datasets, namely CIFAR-10~\cite{cifar}, CelebA~\cite{CelebA}, LSUN-BEDROOM~\cite{lsun} and CelebA-HQ~\cite{celebahq}. The LSUN-BEDROOM dataset~\cite{lsun} contains 3M images which are randomly partitioned into a test set of around 30k images and a training set containing the rest. We use $128\times128\times3$ version of CelebA-HQ with 30k images. We only compare our method with Relativistic GAN and WGANGP on CelebA-HQ due to limited computation resources. 
\\
\textbf{Settings} For CIFAR-10, we use the Resnet~\cite{resnet} architecture proposed in~\cite{generationstability}(with spectral normalization layers removed). For CelebA, LSUN and CelebA-HQ, we used a DCGAN architecture as in~\cite{SNGAN}. We apply Adam optimizer on all experiments as Table~\ref{settings} shows. We used 1 discriminator updates per generator update. The batch size used was 64. Other details of our experiments settings are provided in \textbf{Supplementary}.
  \begin{table*}
   \centering
        \caption{Experiments Settings}
 	\begin{tabular}{rccccc}
		\hline
		\multicolumn{1}{l}{Dataset} & $lr_{g}$& $lr_{d}$    & $\beta_{1}$     & $\beta_{2}$     & Iterations \\
		\hline
		\multicolumn{1}{l}{CIFAR-10} & 0.0002& 0.0001 & 0.9    & 0.999& 600k \\
		\multicolumn{1}{l}{CelebA} & 0.0002& 0.0001 & 0.9    & 0.999& 400k \\
		\multicolumn{1}{l}{LSUN} & 0.0001& 0.0001 & 0      & 0.9 & 400k \\
		\multicolumn{1}{l}{CelebA-HQ} & 0.0001& 0.0001 & 0      & 0.9& 250k \\
		\hline
	\end{tabular}\label{settings}
	\vspace{-4mm}
   \end{table*}
\\
\textbf{Evaluation} To compare the sample quality of different models, we consider three different scores: IS~\cite{IS}, FID~\cite{TTUR} and KID~\cite{KID} which are based on the pre-trained Inception network~\cite{inception} on ImageNet~\cite{imagenet}. 
\\
\begin{figure*}
\begin{tabular}{c@{}c@{}c@{}}
		\includegraphics[width=0.32\linewidth, height=0.32\linewidth]
		{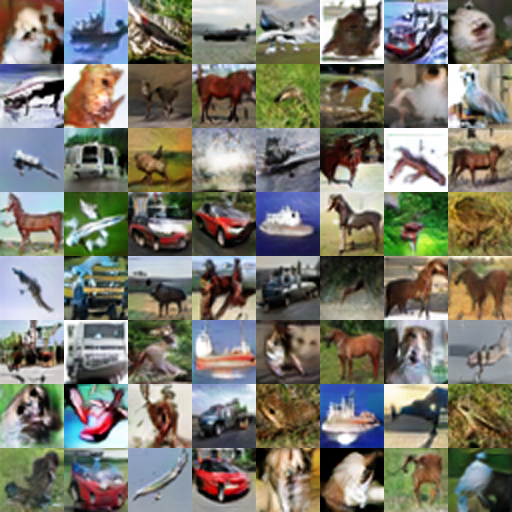} \ 
		&\includegraphics[width=0.32\linewidth, height=0.32\linewidth]
		{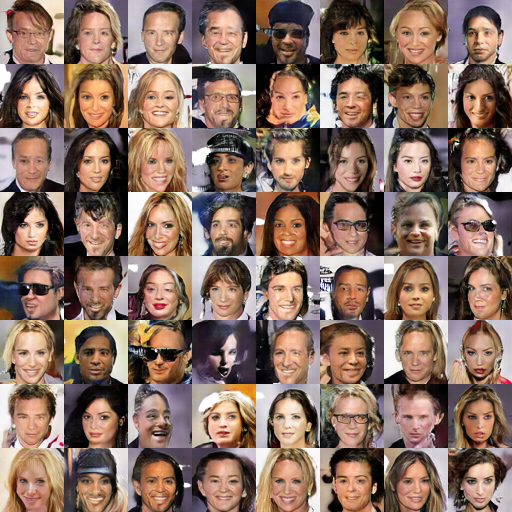}& \
		\includegraphics[width=0.32\linewidth,height=0.32\linewidth]
		{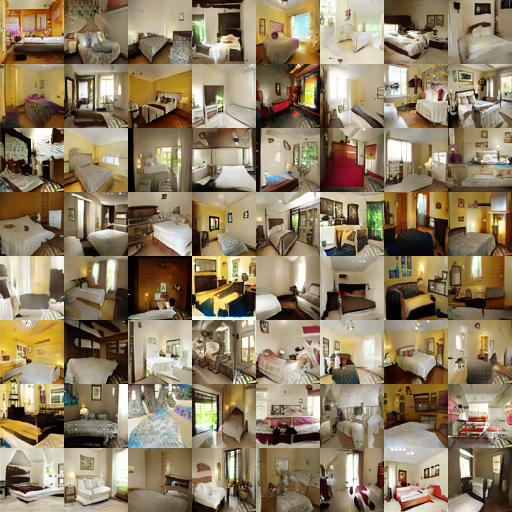} \ 
		\\
        {\footnotesize (a) CIFAR-10} &
		{\footnotesize (b) CelebA}
		&	{\footnotesize (c) LSUN}
	\\
	\end{tabular}
	\caption{Generated Samples(no cherry-picking)} 	 
     \label{fig:Generated}
         \vspace{-3mm}
         \end{figure*}
\textbf{Results and Analysis}
Some random generated samples on 3 data sets are shown in Figure~\ref{fig:Generated}. More generated images and evaluation scores are provided in \textbf{Supplementary 6}. From Table~\ref{scores} we could find  RelatioGAN$^*$ is also highly competitive on single class data sets \emph{i.e.} CelebA, LSUN, while RelationGAN achieves the best performance on CIFAR-10. As we discussed in Sec.\ref{rmmr}, the variant loss of $Ours^*$ is more relaxed and suitable for evenly distributed data sets while the loss of $Ours$ in eq.~\eqref{relu-mean} is more strict and performs better on multi-class or harder data sets (also performs best on Stacked MNIST).

\begin{table*}[ht]
\centering
\setlength{\tabcolsep}{3pt}\label{scores}
	\renewcommand{\arraystretch}{1}
	\caption{The Comparisons of FID, KID Score and IS. RelationGAN represents Relation GAN with objective function in equation~\eqref{relu-mean} and RelationGAN$^*$ represents Relation GAN with objective function in equation~\eqref{mean-relu}. The best two scores are shown in {\color{red}{red}} and {\color{green}{green}}, respectively.}\label{scores}
	\scalebox{0.8}{
\begin{tabular}{l|ccc |ccc }  
\hline
		 & \multicolumn{3}{c|}{CIFAR-10} 
        & \multicolumn{3}{c}{CelebA} 
       \\
		\cline{2-7}
		& FID &KID&IS& FID &KID&IS
        \\
\hline
Vanilla GAN
&26.46$\pm$0.12
&1.88$\pm$0.061
&6.73$\pm$0.081
&34.43$\pm$0.15
&3.01$\pm$0.044
&2.68$\pm$0.020
\\
LS-GAN 
&
\color{green}{14.9$\pm$0.061}
&\color{green}{ 1.31$\pm$0.056}
&\color{red}{7.74$\pm$0.12}
&\color{green}{19.63$\pm$0.11}
&\color{green}{1.84$\pm$0.045} 
&2.5$\pm$0.021
\\
WGAN-GP&63.56$\pm$0.14
&8.01$\pm$0.068
&3.56$\pm$0.038
&66.06$\pm$0.27
&9.06$\pm$0.081
&2.60$\pm$0.029
\\
Relativistic
& 23.96$\pm$0.15
&1.88$\pm$0.061
&0.061$\pm$0.081

&26.71$\pm$0.10
&2.08$\pm$0.050
&\color{green}{3.02$\pm$0.024}
\\
\hline
RelationGAN
&\color{red}{13.52$\pm$0.060}
&\color{red}{1.26$\pm$0.052}
&\color{green}{7.74$\pm$0.18}

&25.37$\pm$0.14
&2.07$\pm$0.044
&2.65$\pm$0.029
\\
RelationGAN$^*$
&47.96$\pm$0.30 
&8.88$\pm$0.072
&3.32$\pm$0.026
&\color{red}{11.99$\pm$0.064}
&\color{red}{1.10$\pm$0.038}
&\color{red}{3.17$\pm$0.036}
\\
\hline
\end{tabular}
}
\scalebox{0.8}{
\begin{tabular}{l|ccc |ccc }  
\hline
		\ 
        &\multicolumn{3}{c|}{LSUN}
       &\multicolumn{2}{c}{CelebA-HQ}
       \\
		\cline{2-7}
		& FID &KID&IS& FID &KID&IS
        \\
\hline
Vanilla GAN
&38.17-$\pm$0.28
&6.61$\pm$0.076
&\color{red}{4.57$\pm$0.010}
&--
&--
&--
\\
LS-GAN &150.61$\pm$0.33
& 21.75$\pm$0.11
&3.57$\pm$0.043
&--
&--
&--
\\
WGAN-GP&\color{green}{14.93$\pm$0.16}
&\color{green}{1.45$\pm$0.042}
&3.77$\pm$0.098
&68.5$\pm$0.19
&7.71$\pm$0.065
&\color{red}{2.31 $\pm$0.017}
\\
Relativistic& 40.84$\pm$0.23
&2.97$\pm$0.045
&4.08$\pm$0.049
&32.24$\pm$0.21
&\color{green}{2.27  $\pm$0.056}
&1.96$\pm$0.038
\\
\hline
RelationGAN&70.24$\pm$0.37
&5.89$\pm$0.078
&\color{green}{4.4$\pm$0.056}
&\color{green}{27.87$\pm$0.17}
&\color{red}{2.21$\pm$0.047}
&2.13$\pm$0.0052
\\
RelationGAN$^*$&\color{red}{12.59$\pm$0.11} 
&\color{red}{1.37$\pm$0.038}
&3.70$\pm$0.081
&\color{red}{26.17$\pm$0.12}
&2.62$\pm$0.043
&\color{green}{2.15$\pm$0.030}
\\\hline
\end{tabular}
}
\vspace{-4mm}
\end{table*}

\subsection{Conditional Image Generation}
We compare the MSGAN~\cite{MSGAN} which is one of the best conditional model on CIFAR-10 datasets. The experiment is applied by simply replace the MS-loss in~\cite{MSGAN} with the relation loss. 
\begin{table}[th] 
\caption{The comparison of FID scores on the CIFAR-10 dataset.}\label{condition-fid}
\centering
\begin{tabular}{p{1cm} p{2cm} p{2cm}} 
\hline
 & MSGAN& RelationGAN \\
\hline
FID &28.73 &\textbf{24.88} \\
\hline
\end{tabular}
\vspace{-4mm}
\end{table}

\subsection{Image Translation}
In addition to image generation task, GANs also gains promising progress in image translation task. We conduct experiments on image style transfer and image super resolution, respectively.

\begin{table}[h!]
\caption{Results of Different Architectures}\label{archi}
    \centering
       \begin{tabular}{l|ccc}
\hline
 & FID(CIFAR-10) \\
\hline
no EM &38.9\\
$0 +3$& 37.37\\
$1 +2$& 28.89\\
$2 +1$& 28.80\\
$3 +0$& 13.52\\
\hline
\end{tabular} 
\vspace{-4mm}
\end{table}

\textbf{Image Style Transfer} For image style transfer task, we adopt the CycleGAN as our baseline model to translate Monet’s painting into photograph. 
FID score is applied to evaluate the quality of generated images. Table~\ref{fid-cycle} shows the comparison of FID scores of generated images. The lower FID represents smaller perceptual difference between target domain images and generated images. We find the both relation loss$^*$ and relation loss performs better than the oigianl adversarial loss in cycle-gan and the reltion loss$^*$ performs best.

\textbf{Image Super Resolution}
For Image Super Resolution task, we employ SRGAN~\cite{srgan} with the relastivistc loss which is the latest proposed loss for gans as our baseline. We denote our baseline as SRGAN$^*$. The train and val datasets are sampled from VOC2012. Train dataset has 16700 images and Val dataset has 425 images. We compare the psrn and ssim on three popular SR datasets: Set5~\cite{set5}, Set14~\cite{set14} and Urban100~\cite{urban}. 

\begin{table}[t]
   \centering
    \caption{The Comparison of FID Scores on Style Transfer Results. The M$\rightarrow$P represents painting to photo,  the P$\rightarrow$M represents photo to painting. RelationGAN and RelationGAN$^*$ represents loss equation~\ref{relu-mean} and equation~\ref{mean-relu}, respectively.}\label{fid-cycle}
     \begin{tabular}{l|ccc|ccc}
\hline
 & FID(M$\rightarrow$P)&FID(P$\rightarrow$M) \\
\hline
CycleGAN &34.00&2.48\\
RelationGAN& \textbf{33.60}&2.26\\
RelationGAN$^*$& 33.71&\textbf{2.21}\\ 
\hline
\end{tabular}
\vspace{-4mm}
\end{table}

Table~\ref{sr-scores} lists the performance of different approaches on five datasets. We can observe that the fid scores of the proposed algorithm perform better than the original method on photo$\rightarrow$painting datasets. 

\begin{table*}
\centering
\caption{Comparison of SRGAN$^*$
and RelationGAN on benchmark data. Highest measures (PSNR (dB), SSIM) in bold. (4$×$ upscaling)}\label{sr-scores}
\begin{tabular}{l|cc |cc |cc}  
\hline
		\ 
        &\multicolumn{2}{c|}{Set5}
       &\multicolumn{2}{c}{Set14}
        &\multicolumn{2}{c}{Urban100}
       \\
		\cline{2-7}
		& psnr&ssim&psnr&ssim&psnr&ssim
        \\
\hline
SRGAN$^*$
&28.40
&0.82
&25.37
&0.73
&23.36
&0.71
\\
RelationGAN &\textbf{28.59}
& \textbf{0.83}
&\textbf{25.52}
&\textbf{0.73}
&\textbf{23.47}
&\textbf{0.72}\\
\hline
\end{tabular}
\vspace{-4mm}
\end{table*}

\subsection{Ablation Study}
We conduct the ablation study on image generation datasets.
We first compare our triplet loss with the loss~\cite{siamese_trip}, whose results are shown in 
Table~\ref{losses}. The formulation of siamese loss function is shown in \textbf{Supplementary 4}. Second, we take a closer look on the impact of our embedding module and relation module. The ``$n+m$'' in the Table~\ref{archi} represents different architectures of discriminator, where the embedding module contains $n$ res-block and the relation module contains $m$ res-block. The ``(0+3)'' represents the samples are concatenated together after first conv-layer and then put into the relation module (RM) which contains 3 res-block. The ``no EM'' represents the samples in which the paired input are packed in the beginning of the discriminator as~\cite{PacGAN}. All experiments are conducted on CIFAR-10.

\textbf{Results and Analysis}
From Table~\ref{losses}, we could find the results of the proposed triplet loss is much better than Siamese loss. The ``-'' represents model collapse in training process. The results in Table~\ref{archi} shows the bigger size of EM could enhance the performance which also demonstrates the effectiveness of our embedding strategy.
\begin{table}[h!]
 \caption{Ablation Losses}
    \label{losses}
    \centering
     \begin{tabular}{l|ccccc}
\hline
 & CIFAR-10 & CelebA\\
\hline
Triplet& 13.42 & 11.9 \\
Siamese&-- & 107.3 \\
\hline
\end{tabular}\label{losses}
\vspace{-4mm}
\end{table}


\section{Conclusion}
In this paper we propose the Relation GANs. A relation network architecture is designed and used as the discriminator, which is trained to determine whether a paired input samples are from the same distribution or not. The generator is jointly trained with the discriminator to confuse its decision using triplet losses.
\\
Mathematically, we prove that the optimal discriminator based on the relation network is a divergence, indicating the distance of generated data distribution and the real data distribution becomes progressively smaller during the training process.
We also prove the generated data distribution will converge to the real data distribution when getting to the Nash equilibrium. 
In addition, we analysis our method and several other GANs in dynamic system. We demonstrate our GAN has excellent convergence by analyzing the dynamic system of the Dirac GANs.
\\
The results of experiments verify the effectiveness of Relation GAN, especially in addressing the mode collapse problem.
Our Relation GAN not only achieves state-of-the-art performance on unconditional and conditional image generation task with the basic architecture and training settings,  
but also achieves promising results in image translation tasks compared with other losses.


\bibliographystyle{splncs04}
\bibliography{egbib}
\end{document}